# Stock Market Price Prediction: A Hybrid LSTM and Sequential Self-Attention based Approach

Karan Pardeshi[1], Sukhpal Singh Gill[1], Ahmed M. Abdelmoniem[1]

[1]School of Electronic Engineering and Computer Science, Queen Mary University of London, UK

**ABSTRACT**— One of the most enticing research areas is the stock market, and projecting stock prices may help investors profit by making the best decisions at the correct time. Deep learning strategies have emerged as a critical technique in the field of the financial market. The stock market is impacted due to two aspects, one is the geo-political, social and global events on the bases of which the price trends could be affected. Meanwhile, the second aspect purely focuses on historical price trends and seasonality, allowing us to forecast stock prices. In this paper, our aim is to focus on the second aspect and build a model that predicts future prices with minimal errors. In order to provide better prediction results of stock price, we propose a new model named Long Short-Term Memory (LSTM) with Sequential Self-Attention Mechanism (LSTM-SSAM). Finally, we conduct extensive experiments on the three stock datasets: SBIN, HDFCBANK, and BANKBARODA. The experimental results prove the effectiveness and feasibility of the proposed model compared to existing models. The experimental findings demonstrate that the root-mean-squared error (RMSE), and R-square ($R^2$) evaluation indicators are giving the best results.

## I. INTRODUCTION

Utilizing historical stock prices within a certain time period is the most crucial step for investors and businesses to forecast future sales and any potential negative earnings. Many researchers and investors who desire to speculate within the stock exchange are lured to that since its fundamentals include high risk and great returns [1]. In the past, the main methods used for determining the value of a developing business were to estimate the profits and incomes and then apply the appropriate cash flow discount rate to those predictions. However, an organization may use this conventional method for forecasting earnings provided that it has enough data on performance, positive earnings, or comparable enterprises. We try and solve this issue by giving financial managers and investors the information and algorithms they require to create wise business decisions.

Researchers have tried to anticipate the longer-term condition of the exchange using models that may analyze historical exchange data due to the supply of both past and present securities market data. This historical data typically includes Date, Open, High, Low, Close, Adjusted Close Prices and Volume as shown in Figure 1. Using some statistical analysis approaches, daily data could be examined and the exchange's future condition might be predicted. The exchange data is additionally naturally noisy and unpredictable because of non-economic variables like natural disasters and political actions. The shortcoming of not accurately capturing the connection between future and past prices is another factor contributing to the unpredictability of the stock data. It would be difficult to forecast the longer-term price of a corporation due to the insufficient information surrounding the stock exchange data, which is often described as noisy features as explained by the researcher [2,3]. The necessity for suitable algorithms and techniques to attenuate risks and maximize profits grew together with the fast expansion of commerce and investment.

|  | Date | Open | High | Low | Close | Adj Close | Volume |
|---|---|---|---|---|---|---|---|
| 0 | 2012-05-02 | 214.600006 | 216.740005 | 212.100006 | 213.945007 | 191.394974 | 2968820 |
| 1 | 2012-05-03 | 213.100006 | 213.100006 | 207.660004 | 208.520004 | 186.541763 | 5035860 |
| 2 | 2012-05-04 | 206.410004 | 206.945007 | 198.500000 | 199.360001 | 178.347244 | 5911990 |
| 3 | 2012-05-07 | 193.600006 | 203.755005 | 188.100006 | 202.615005 | 181.259155 | 9728110 |
| 4 | 2012-05-08 | 204.000000 | 204.949997 | 193.000000 | 195.895004 | 175.247467 | 8083050 |
| ... | ... | ... | ... | ... | ... | ... | ... |
| 2479 | 2022-05-24 | 462.000000 | 466.700012 | 460.200012 | 462.350006 | 455.250031 | 449676 |
| 2480 | 2022-05-25 | 461.399994 | 464.549988 | 452.149994 | 454.200012 | 454.200012 | 473896 |
| 2481 | 2022-05-26 | 457.399994 | 470.000000 | 452.549988 | 469.000000 | 469.000000 | 916936 |
| 2482 | 2022-05-27 | 471.950012 | 474.950012 | 467.149994 | 469.000000 | 469.000000 | 817811 |
| 2483 | 2022-05-30 | 473.799988 | 476.700012 | 471.200012 | 474.450012 | 474.450012 | 533868 |

*Figure 1: SBIN stock data snapshot from May 2012 to May 2022.*

This paper tackles these difficulties in predicting stock prices. The target is to attenuate risks for investors within the stock exchange and decision-makers within the financial sector while capturing difficult and constantly changing pricing more promptly. Using the State Bank of India (SBIN) historical stock data [1] as an example, our study compares several statistical forecasting techniques including Bidirectional Long short-term memory (BiLSTM), Convolutional Neural Network (CNN), Facebook Prophet, LSTM-CNN, Auto Regressive Integrated Moving Average (ARIMA) models to exactly predict the longer-term values of the stock [24]. The prediction results of the proposed LSTM-SSAM model track accurately the daily 'adjusted closing' prices of the SBIN stock market index which dramatically raises the bar for developing effective business plans. And, we use certain significant technical indicators just like the Simple Moving Average (SMA) as supplementary forecasts so as to extend the robustness of our proposed approach.

### A. Motivation

The main motivation for trying to predict stock market prices is the potential return on investment. Even a fraction of a second's knowledge of a stock's price can result in large earnings. Researchers and investors are continually looking for the stock's future price that will provide them with the highest profits. You can add a level of certainty by trying to understand the historical trend of a particular stock which is available on the Internet.

To minimize losses and maximize profits, our approach can be very helpful in increasing the accuracy of stock market price predictions. The prediction techniques we outline here are up to date and are deemed to be extremely helpful in identifying prospective buying and selling opportunities for the stock of interest. This allows investors to choose when to invest their money to get more returns.

### B. Contributions

Our contributions can be summarized as follows:
1. We study and analyze the problem of stock prediction and present the prominent existing methods used for this purpose.
2. We propose a Hybrid LSTM and Self-Attention based approach to tackle this problem which presents a novel solution to the problem.
3. Using common stock datasets, we evaluate the proposed approach and compare them with the existing methods to show its effectiveness.
4. Our results show that the proposed approach shows superior performance and can accurately predict future stock prices.

The outline of this paper is as follows. Section 2 provides the related work; Section 3 shows the methodology; Section 4 provides the analysis; Section 5 is the experimental results followed by conclusions and future directions in Section 6; and finally Section 7 provides the references in this work.

## II. RELATED WORK

### A. *Long-Short Term Memory (LSTM)*

LSTMs can analyze both individual data points (images) and whole data sequences (videos, time series, etc.). LSTM is a type of Recurrent Neural Network (RNN), which was first introduced recently, and has gained appeal among financial specialists [4]. Similar to conventional RNN frameworks, the LSTM model may learn data over one or more layers [5]. LSTM networks were developed to address the exploding and vanishing gradient problems that can occur when training traditional recurrent neural networks, so they are appropriate for making predictions based on time series data. This is because there may be lags of unknown length between significant events in a time series. In contrast to LSTM, machine learning requires a lot of time.

One of the researchers used the LSTM model to forecast the values of S&P 500 equities from 1992 to 2015 [6]. He used three years of data to train the LSTM model and the following year to evaluate it. In his study, the model beat the logistic regression classifier, random forest [6]. On the other hand, a study using the LSTM model was done on the Chinese Stock Market to anticipate prices [7]. The LSTM model, an algorithmic approach for interpreting time series data, assumes that the underlying process is unknown. The LSTM model was more accurate in predicting stock price variations than the CART (Classification and Regression Tree) model in the study conducted by [8]. This is because the LSTM model has the capacity to use deep learning to assess sequential input, extract useful information, and discard irrelevant information. When evaluating the LSTM model, the work only considered the basic LSTM model [8].

### B. *Bidirectional Long Short-Term Memory (BiLSTM)*

Although LSTM is capable of gathering information about long-distance features, it only makes use of data that is accessible before output time. When creating predictions for time series, it is essential to fully take into consideration the forward and backward information law of the data, since this may greatly improve forecast accuracy. The two halves of BiLSTM are forward and backward LSTM. BiLSTM may utilize the past and future data to draw conclusions that are more extensive and accurate because it considers the changing laws of the data both before and after data transfer, as opposed to the one-way-state transmission in the ordinary LSTM. It has shown exceptional performance and some work used it in applications for water level prediction [9].

*C. Convolutional Neural Network (CNN)*

CNN is used in a variety of applications, including vision, time series analysis, and the extraction of predictive characteristics from data [9]. The traditional CNN has layers for convolution, activation, and pooling [10]. An effective convolution method is used to abstract the original input while representing data characteristics in a higher-level form. The majority of CNN is made up of both the pooling layer and the convolution layer. Each convolution layer consists of several convolution kernels, which can be calculated as shown in Equation (1).

$$l_t = \tanh(x_t * k_t + b_t) \quad \quad \quad (1)$$

Where:
- $l_t$ is the output value after convolution,
- tanh is the activation function,
- $x_t$ is the input vector,
- $k_t$ is the weight of the convolution kernel, and
- $b_t$ is the bias of the convolution kernel.

After the convolution operation of the convolution layer, the features of the data are retrieved, however, the extracted feature dimensions are quite high. In order to solve this problem and reduce the cost of network training, a pooling layer is added after the convolution layer to reduce the feature dimension.

*D. Auto-Regressive Integrated Moving Average (ARIMA)*

Originally introduced and developed by Box and Jenkins (1976), the general model incorporates the AR (autoregressive), I (differencing), and MA (moving average) components. It involves a variety of stages for finding, developing, and assessing ARIMA models using time series data [14]. It is sometimes known as the Box-Jenkins technique. One of the approaches for financial forecasting is most frequently used in this model [15, 16, 17]. For short-term forecasting, ARIMA models have proven to be helpful. It consistently outperformed complicated structural models in terms of short-term prediction [13]. ARIMA has 3 main parameters namely p, d, q, where:
- p is a kind of order of autoregressive part for correlation within itself.
- d is the number of times that raw observations are contrasted, this is considered as 'I' - Integrated with ARIMA.
- q is the moving average parameter.

*E. FaceBook Prophet Tool*

Autoregressive models are among the most frequently used to forecast future predictions. Briefly, the output variable is linearly dependent on both its own prior values and a stochastic component in the autoregressive model (an imperfectly predictable term). The Prophet model was recently proposed by Facebook in an effort to create a model that could identify seasonality in time-series data [11]. This model is openly accessible. By using additive regression models, Prophet can track seasonality on a daily, monthly, and annual basis as well as the effects of holidays. For trend forecasting, Prophet uses a piecewise linear model. The mathematical Equation (2) for the Prophet model is as follows:

$$y(t) = g(t) + s(t) + h(t) + e(t) \quad \quad \quad (2)$$

Where:
- g(t) denotes the trend;
- S(t) denotes recurring changes (weekly, monthly, or yearly);
- H(t) is a symbol for the impacts of holidays (recall: Holidays impact businesses); and
- e(t) is the error term.

Even when dealing with thousands of observations, prophet model fitting procedures are typically faster and don't require any data pre-processing. Missing data and outliers are also addressed [11]. In Facebook Prophet, both linear and nonlinear time functions are supported. In exponential smoothing, a similar method of modelling seasonality as an additive component is applied. This library is so important that it can address problems with seasonality and stationary properties in data. There are a few restrictions with Facebook Prophet, such as the need for input columns with the Identifiers "ds" and "y," where "ds" stands for Date and "y" for the target variable. In this scenario, a trend might be rising or falling, positive or negative [12].

### III. METHODOLOGY

In this section, we discuss the methodology towards our proposed model and approach for stock price prediction.

*A. Dataset*

The stock used in this study is in form of historical data of the SBIN stock data which includes "open, low, high, close, and adjusted close prices and trading volume". The daily price data used, which spans a period of 10 years and includes 2484 observations, is taken from Yahoo Finance [1]. We consider the "Adjusted Close" price; because the closing price refers to the cost of share at the end of the day, while the adjusted closing price accounts for items like dividends, stock splits, and new stock offerings. It might be considered that the adjusted closing price is a more accurate indicator of the stock value since it starts where the closing price ends. To study the general trend of the stock, we have divided the dates into months and by observing the graph shown in Figure 2, we can say that from November till February the stock price of SBIN is high due to festivals and year-end closing.

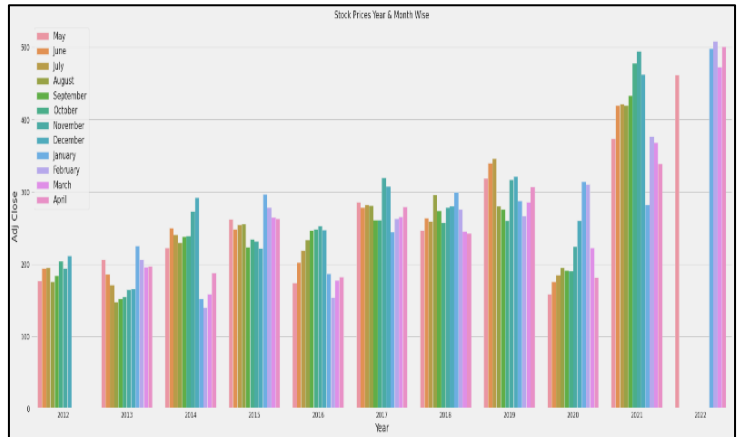

*Figure 2: General trend of SBI stock market month and year wise.*

### B. Feature Engineering

The original data of the stock has the following fields:
- Date: Trading Date.
- Open: The price at which the stock is initially traded.
- High: The day's peak trading price.
- Low: Lowest price during the trading day.
- Close: The final price at which the stock changed hands throughout the trading day.
- Adjusted Close: The price changes in response to corporate activities on the closing price.
- Volume: The total number of shares that were traded throughout the day.

For our model, we concentrate on 'Date' and 'Adjusted Close' prices. An adjustment to the computation is made to the closing price of a stock, resulting in the adjusted closing price. Compared to the closing price, it is more detailed and precise. Because external circumstances might have caused changes in the genuine price, the adjustment made to the closing price represents the stock's true price. To improve the accuracy of the final data forecasts, several methods and algorithms are used. Data analysis may be enhanced by using simple moving averages, and RMSE and R2 scores.

### C. Correlation Analysis

A table that shows the correlation coefficients between several variables is called a correlation matrix. The correlation between any two variables is displayed in each column of the table as shown in Figure 3. The values are indicating how each column relates to the other column. The table demonstrates the strong correlation between practically all the fields in the dataset, except for the volume.

|           | Open     | High     | Low      | Close    | Adj Close | Volume    |
|-----------|----------|----------|----------|----------|-----------|-----------|
| Open      | 1.000000 | 0.999025 | 0.999015 | 0.997905 | 0.996717  | -0.386017 |
| High      | 0.999025 | 1.000000 | 0.998690 | 0.999163 | 0.998049  | -0.370782 |
| Low       | 0.999015 | 0.998690 | 1.000000 | 0.999030 | 0.997647  | -0.392613 |
| Close     | 0.997905 | 0.999163 | 0.999030 | 1.000000 | 0.998720  | -0.378548 |
| Adj Close | 0.996717 | 0.998049 | 0.997647 | 0.998720 | 1.000000  | -0.390092 |
| Volume    | -0.386017| -0.370782| -0.392613| -0.378548| -0.390092 | 1.000000  |

*Figure 3: Correlation Values for SBIN stock data.*

### D. Simple Moving Average (SMA)

It calculates the average value of the previous 'n' data points. It is a useful tool that may help determine if a stock price trend will continue or reverse. Equation (3) shows how SMA is calculated as follows:

$$\frac{(A_1 + A_2 + \cdots + A_n)}{n} \quad \ldots\ldots\ldots\ldots\ldots\ldots\ldots\ldots\ldots\ldots\ldots\ldots\ldots\ldots\ldots\ldots\ldots\ldots\ldots\ldots\ldots \quad (3)$$

where,
  (A1, A2, …, An) are the prices; and
  n is the number of total periods.

The number of periods in that range is used to calculate the percentage of an elected assortment of prices, or routinely closing prices. A simple moving average lowers volatility and makes it easier to identify a stock's price trend. A rising simple moving average indicates that the stock price is also rising. If it is heading downward, the price of the stock is falling which can be viewed in Figure 4. The smoother the simple moving average, the longer the time period for the moving average.

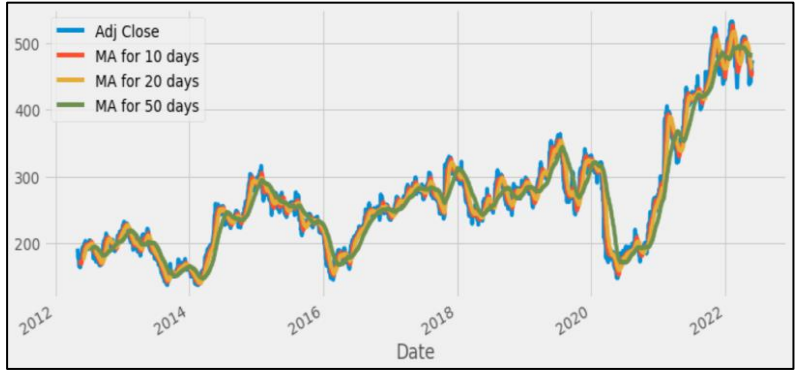

*Figure 4: Calculating Moving Average for 10,20,50 days.*

### E. Root Mean Square Error (RMSE)

RMSE is used in this instance to determine the most appropriate moving average for the data. A low RMSE value represents low error. In climatology, forecasting, and regression analysis, root means the square error is frequently used to validate experimental results. Equation (4) for RMSE is obtained as follows:

$$RMSE = \sqrt{\frac{1}{n}\sum_{i=1}^{n}(f_i - o_i)^2} \quad \ldots (4)$$

where,
  $\Sigma$ is the summation of all values,
  f is the predicted value,
  o is observed or actual value,
  $(f_i - o_i)^2$ is the difference between predicted and observed values and squared,
  N is the total sample size.

## IV. ANALYSIS

### A. Experimental Design

The study is divided into six sections: data collection, data preprocessing, model training, model saving, model testing, and prediction findings. Figure 5 displays the conceptual flowchart for the method analysis and design.

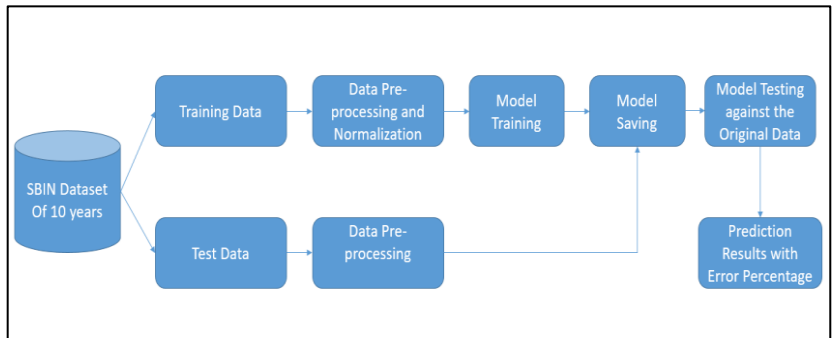

*Figure 5: Illustration of the Experimental process.*

Since the original yahoo finance dataset cannot be used directly for model training and testing, data preprocessing is necessary. The issue of inconsistent data magnitude is then addressed by using data normalization. Finally, a useful dataset for the experiment may be created. Data normalization entails transforming the data into a specified interval and scaling the data to a particular scale. In this experiment, the value of the data is changed to, which increases the model's accuracy and convergence speed. Equation (5) of the transformation function is as follow:

$$x^* = \frac{x - min}{max - min} \quad \quad \quad \quad \quad \quad \quad \quad \quad \quad \quad \quad \quad \quad \quad \quad \quad \quad (5)$$

where max and min stand for the sample data's maximum and minimum values, respectively.

B. *Experimenting with other algorithms and configurations*

The algorithms used for time series are LSTM, BiLSTM, CNN, FB Prophet, ARIMA, and hybrid algorithms. Our study finds that LSTM is very effective and it can more accurately forecast future values and capture previous trend patterns successfully. The configurations of the algorithms are listed in Table 1 shown below.

*Table 1: Configuration of other algorithms.*

| Algorithms | Configuration |
| --- | --- |
| LSTM (Unit 1) | The number of hidden units 1 |
| LSTM (Unit 50) | The number of hidden units 50 |
| CNN + BiLSTM | Conv1d + MaxPooling +BiLSTM units 50 + Dropout + dense |
| LSTM + CNN | LSTM unit 50 + Conv1d + MaxPooling + Flatten + Dense *2 |
| Facebook Prophet | Trends of Seasonality |
| ARIMA | p,d,q(0,1,0) |

C. *The Proposed Algorithm - LSTM with Sequential Self-Attention Mechanism (LSTM-SSAM)*

In the following, we will discuss the key components of the proposed LSTM-SSAM approach.

C.1 **Long Short-Term Memory**

Using Google's open-source Tensor Flow software, the LSTM model was developed in Python using Keras, a high level neural networks API [18]. It uses the aforementioned dataset, which consists of SBIN data points for the time periods 2012-05-02 through 2022-05-31, and splits it into training and test data in proportions of 90% and 10%, respectively. The output (Y) is a forecast of the "Adj Closing" price for the upcoming day using the selected feature(s) of the present and previous number of days (equal to the time step) as input (X).

Our model is built using a sequential model with one LSTM layer and one dense layer. We have set LSTM return sequences to be True value and the LSTM utilizes 50 units (dimensionality of the output space). When the return sequences argument is set to True, all of the time steps' hidden states are output. The array's content consists of all the hidden states for each LSTM layer time step. For the purposes of this study, different parameter settings have been developed by trial and error. As a result, many parameter choices were attempted and tested by repeatedly running the model, and values for the hyperparameters that balance the trade-offs between time efficiency and performance are chosen.

In the selection process, several settings that reduced the difference between high and low values were tested. A value of **10** was the best batch size, both in terms of performance and efficiency, after various batch sizes were tested (i.e., the number of samples per gradient update). A value of **50** was chosen as the number of times to loop through the whole dataset because epoch sizes above 50 showed little to no improvements. The mean squared error (MSE) is applied to the loss function. Adam optimizer was chosen among the different optimizers available as Adam gave us the best results. Adam optimizer, which iteratively changes network weights in training data, performed better than Stochastic Gradient Decent (SGD). On the other hand, SGD keeps a constant learning rate during all weight changes. Adam was used because empirical data showed that it is effective in practice and even compares favourably to other stochastic optimization techniques.

The architecture of the LSTM is shown in Figure 6. A gradient may transfer from one hidden layer to the next without decreasing due to the memory of LSTM units, which also tackles the problem of vanishing gradients. The cells are a representation of the transport line that runs through each cell (the top line in Figure 6), connecting data from previous modules and the current module. It has been established that LSTMs are more powerful than standard RNNs. The LSTM layer is made up of four neural network layers that interact in a certain way.

A typical LSTM device consists of three distinct parts: **a cell, an escape door, and a forgotten door**. Cells' main responsibilities are to categorize values over a range of arbitrary periods and to keep track of the data flow in and out of them. In order to regulate the state of each cell, each LSTM needs three different kinds of doors: Between 0 and 1, Forget Gate generates a number, with 1 denoting "to keep that completely" and 0 denoting "to ignore it altogether." Memory Gate selects the most recent data that needs to be processed in the cell. The values that are modified are then determined by a sigmoid layer called the "input layer." After that, a "tanh" layer generates a new candidate value vector that might be added to the state. Each cell's output is determined by the gate output.

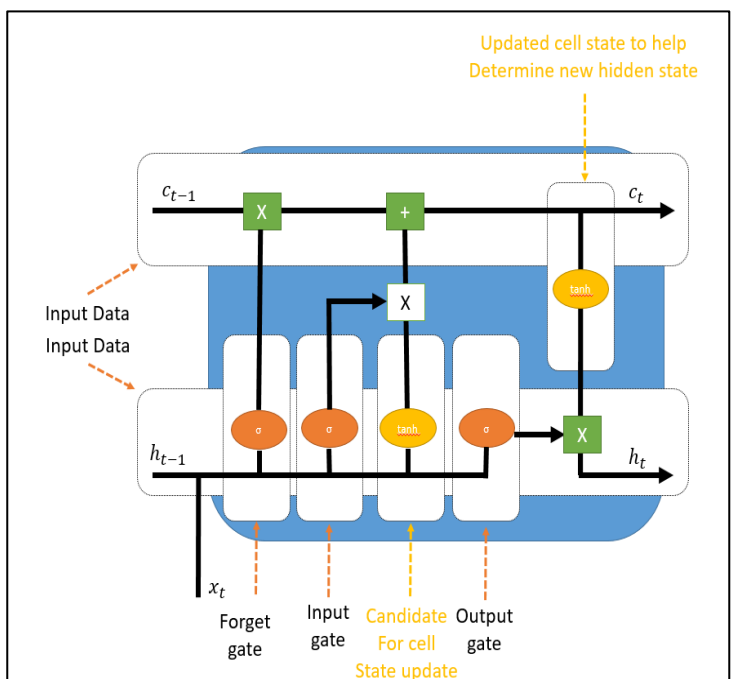

*Figure 6: Architecture of Long Short Term Memory.*

An LSTM will have three separate dependencies depending on the data it receives:
- The prior condition of the cell (specifically, the data that remained in the memory following the previous time step).
- The previous hidden state (This is the same as what the previous cell's output was, for example).
- The input for the current time step (i.e. the most recent data being input at that moment).

These dependencies may be explained as follows in relation to the field of financial asset price prediction:
(1) **The Previous Cell State** – The stock's trend of the previous day.
(2) **The Previous Hidden State** – The stock's price of the previous day.
(3) **The Input at the Current Time Step** – Other elements that may have an impact on price. These can be daily investment news alerts that are released publicly.

The horizontal line **'ct'** in the diagram shown in Figure 6 stands for the cell state, which symbolizes the LSTM's memory. This is consistent with the stock's previous day's pattern, as we immediately noted above (1). This data will be processed by the cell, which will also handle any other data that comes in. In our stock prediction scenario, the data from earlier time steps, in particular, the stock price from the previous day, will be present in the line denoted by **'ct-1'**, which is the Hidden State stated above (2). The horizontal line shown by **'ht-1'** represents the input during this time, which is the stock price at the present moment as stated above (3). The LSTM will provide a result based on information from prior stock prices (Hidden State), the current price in combination with the trend from the previous day (Cell State), and so on.

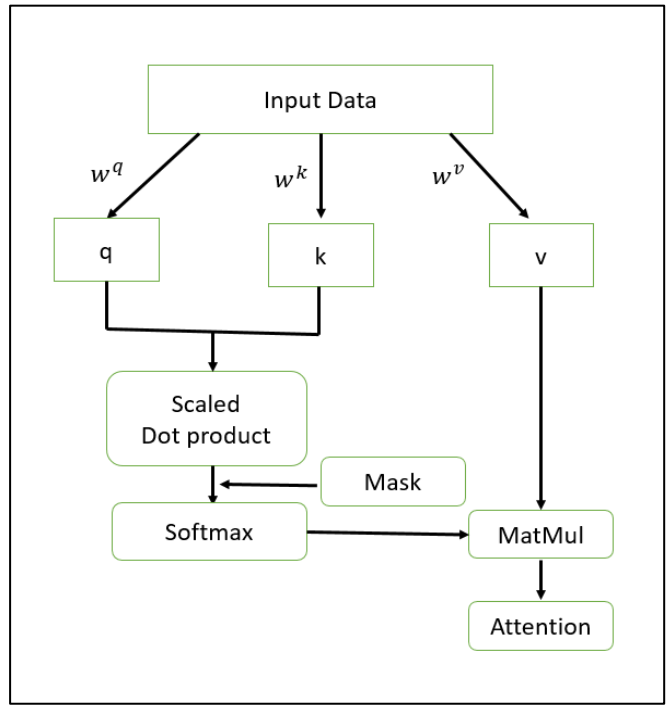

*Figure 7: Self Attention Mechanism Architecture.*

**C.2 Self-Attention Mechanism**
The self-attention model allows inputs to interact with one another (i.e., calculate the attention of all other inputs with regard to one input) as illustrated in Figure 7. While producing output, the attention mechanism enables output to concentrate attention on input. The first phase involves multiplying each of the encoder input vectors by each of the three weight matrices (W(Q), W(K), and W(V)) that we trained throughout the training process [20–21]. For each of the input vectors, this matrix multiplication will produce three vectors: t**he key vector, the query vector, and the value vector**.

The second phase of calculating self-attention is the Query vector from the current input is multiplied by the key vectors from other inputs. The third phase involves multiplying the score by the square root of the dimensions of the key vector (dk). The reason for this is that when we use the softmax function in the future, certain self-attention scores end up being extremely low. In fourth phase all of the self-attention scores will be subjected to the softmax function. In the fifth phase, the value vector is multiplied by the determined vector from the previous step. In the last phase, the weighted value vectors obtained in the stage before are added to provide the self-attention output.

All input sequences are subjected to the aforementioned method [19]. The self-attention matrix for the input matrices (Q, K, and V) is determined mathematically by Equation (6):

$$\text{Attention}(Q, K, V) = \text{softmax}\left(\frac{QK^T}{\sqrt{d_k}}\right)V \quad \dots \quad (6)$$

where the vectors for the query, key, and value are combined together as Q, K, and V.
- **Q - Query**: Queries are a collection of vectors that you obtain by merging the input vector with the query weights (Wq), and you wish to calculate attention for these vectors.
- **K - Key**: keys are a collection of vectors that are produced when you combine the input vector with the key weights; these are the vectors that you will use to measure attention.
- **V - Value**: When you combine an input vector with value weights, you obtain a collection of vectors called values. These vectors are then multiplied by the softmax attention probabilities. Value vectors enable the selection of pertinent terms.

Rectified linear unit (ReLU) function is the best activation function to utilize for forecasting individual stock returns since it is most frequently used in the literature and empirical investigations [22]. ReLU allows non-linear properties in the data, giving us more alternatives with our dataset. The ReLU function is described as follows in Equation (7):

$$\text{ReLU}(x) = \begin{cases} 0 & \text{if } x < 0 \\ x & \text{otherwise} \end{cases} \quad \dots \quad (7)$$

As a result, the ReLU activation function only requires the total weight inputs from each layer; in the absence of adequate data, it outputs zero; in the presence of sufficient data, it receives an x value. For the specific issue we are dealing with, we use this as the last layer since the linear activation function is always defined as the last layer output for regression issues. After building the network structure and the relative activation functions, it is essential to provide a measure to assess the model's performance in order to increase prediction accuracy.

There are a number of hyperparameters that must be determined, including the number of flattened and dense neurons in each layer, as shown in Table 2.

*Table 2: Parameters setting of the proposed method*

| Layer | Output Shape | Parameters |
|---|---|---|
| LSTM (Unit 50) | (None, 10, 50) | 10400 |
| Seq_self_attention | (None, 10, 50) | 3265 |
| Flatten | (None, 500) | 0 |
| Dense | (none, 1) | 501 |

## V. EXPERIMENTAL RESULTS

In order to verify the effectiveness of the proposed model, different models including LSTM (unit 1), LSTM (Unit 50), CNN-BiLSTM, LSTM-CNN are compared on the SBIN stock dataset collected for the project. The predicted results are shown in Figure 8 which presents on the x-axis the 'Date' and on the y-axis the 'Adj Closed' price. As shown in this figure, the red line is the test data of the 'Adj Close" price of the stock which is the actual value, and the other lines are the predicted values of the 'Adj Closed' price.

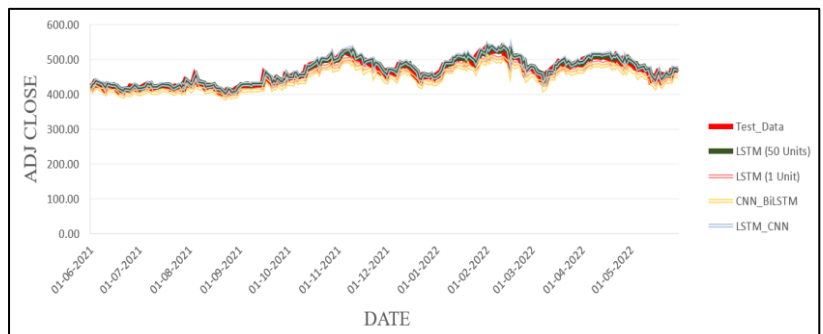

*Figure 8: SBIN graphical representation of other algorithms.*

The prediction errors of different models on the SBIN stock dataset is listed in Table 3. As shown in this table, LSTM (unit 50) have higher predictive power than others.

*Table 3: Comparison of Algorithms on SBIN.*

| ALGORITHM | RMSE | Time(Sec) | R2 Score |
|---|---|---|---|
| LSTM (Unit 1) | 14.37 | 144.0 | 0.83 |
| **LSTM (Unit50)** | **10.80** | **144.1** | **0.905** |
| CNN+BiLSTM | 21.32 | 144.0 | 0.63 |
| LSTM+CNN | 19.78 | 144.2 | 0.67 |
| FB_Prophet | 88..64 | 7.4 | -0.08 |
| ARIMA | 49.82 | - | -0.09 |

We can take LSTM (unit 50) and agument it with the self-attention mechanism to enhance the performance of the LSTM mode. We present the results in Table 4 which verify the effectiveness of the LSTM-SSAM model. It can achieve superior performance compared to the LSTM model over all the metrics (i.e., RMSE, time and R2 score). And, the values of actual data and predicted data are in the same range and the error difference is lower compared to other algorithms. Figure 9 shows the red line in the same range of Test_data (actual price from June 2021 to May 2022).

*Table 4: Comparison of LSTM (unit 50) and Proposed Model.*

| ALGORITHM | RMSE | Time(Sec) | R2 Score |
|---|---|---|---|
| LSTM(Unit50) | 10.80 | 144.1 | 0.905 |
| **Proposed Model** | **8.17** | **108.2** | **0.946** |

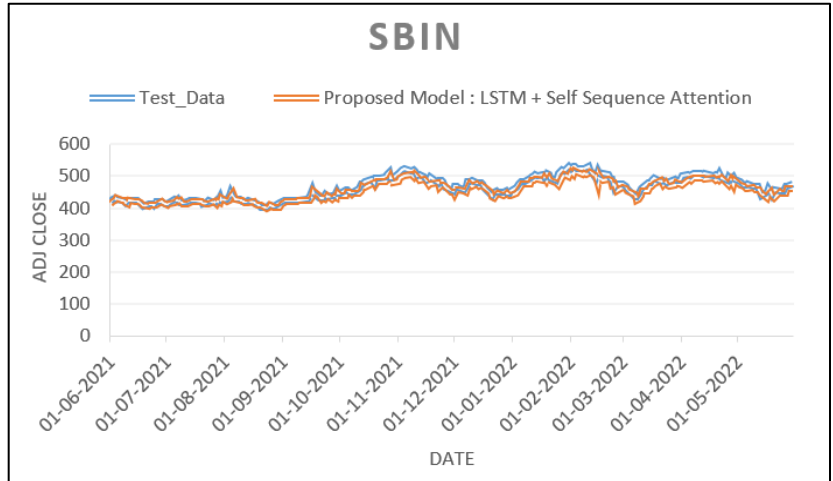

*Figure 9: Graph of Proposed Model predicting the values of SBIN Test dataset from June 2021 to May 2022.*

Table 5 gives the values for the forecast error between the actual price and the predicted price and its error percentage. The results show that the error is significantly low and its percentage is less than ~3%. The forecast error is calculated by subtracting the actual values and the predicted values as shown in Equation (8):

Forecast Error = [Actual Price] – [Predicted Price] …………………………………………………… (8)

Similarly, the forecast percentage is calculated by dividing forecast error and actual price multiplied by 100 as in Equation (9):

$Forecast\ Error\ (\%) = Abs\left(\frac{[Forecast\ Error]}{[Actual]}\right) \times 100$ …………………………………………………… (9)

Where Abs is the absolute value of the number.

*Table 5: Proposed model predictions from June 2021 to May 2022.*

| Date | Test_Data | Predicted | Forecast Error | Error Percentage |
|---|---|---|---|---|
| 01-06-2021 | 422.060 | 414.115 | 7.945 | 1.882 |
| 02-06-2021 | 426.548 | 421.643 | 4.904 | 1.150 |
| 03-06-2021 | 432.899 | 425.945 | 6.953 | 1.606 |
| 04-06-2021 | 426.942 | 432.451 | -5.510 | 1.291 |
| 07-06-2021 | 425.760 | 427.028 | -1.268 | 0.298 |
| 08-06-2021 | 420.591 | 426.185 | -5.595 | 1.330 |
| 09-06-2021 | 414.978 | 421.085 | -6.107 | 1.472 |
| 23-05-2022 | 453.872 | 454.919 | -1.047 | 0.231 |
| 24-05-2022 | 455.250 | 452.759 | 2.491 | 0.547 |
| 25-05-2022 | 454.200 | 454.959 | -0.759 | 0.167 |
| 26-05-2022 | 469.000 | 453.817 | 15.183 | 3.237 |
| 27-05-2022 | 469.000 | 468.350 | 0.650 | 0.139 |
| 30-05-2022 | 474.450 | 467.755 | 6.695 | 1.411 |

We have also compared the same algorithms with 2 different stock data namely HDFC BANK and BANKBARODA datasets. Table 6 shows the results of testing the various in this study models and we observe that the proposed model outperforms the other models and has the best accuracy. Here also we can observe that LSTM (unit 50) has greater accuracy than other algorithms and our proposed model LSTM-SSAM model outperforms the LSTM mode with accuracies of **90.74%** for HDFCBANK and **96.08%** for BANKBARODA.

*Table 6: Results for (HDFCBANK and BANKBARODA) stocks.*

| ALGORITHM | HDFC RMSE | R2 Score | BANK BARODA RMSE | R2 Score |
|---|---|---|---|---|
| LSTM(Unit 1) | 28.07 | 87.90 | 3.51 | 91.50 |
| LSTM(Unit50) | 27.65 | 88.26 | 3.13 | 93.25 |
| CNN+BiLSTM | 39.51 | 76.04 | 2.67 | 95.08 |
| LSTM+CNN | 44.26 | 69.99 | 2.70 | 94.95 |
| FB_Prophet | 524.80 | -6.39 | 126.41 | -0.99 |
| ARIMA | 124.53 | -0.90 | 17.71 | -0.001 |
| **Proposed Model** | **24.55** | **90.74** | **2.38** | **96.08** |

Finally, the graph in Figure 10 and Figure 11 compares the Test_Data and the proposed model predicted price. It shows that the proposed model has the least error and demonstrates its effectiveness in the future price of these two stocks.

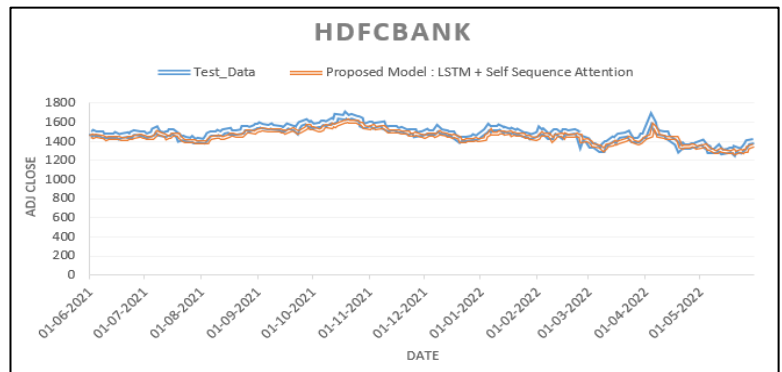

*Figure 10. Graph of Proposed Model predicting the values of HDFC Test dataset from June 2021 to May 2022.*

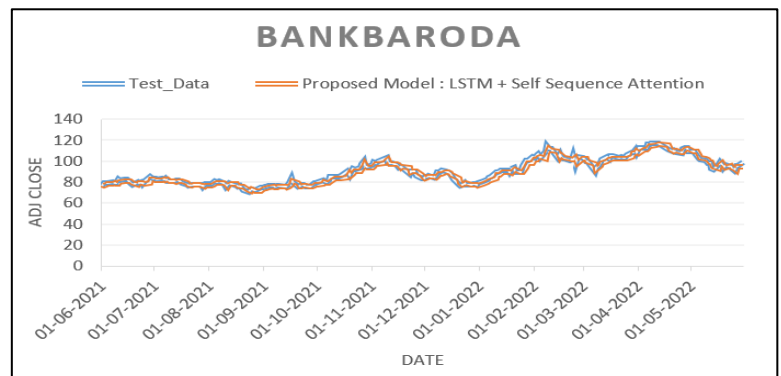

*Figure 11. Graph of Proposed Model predicting the values of BANKBARODA Test dataset from June 2021 to May 2022.*

In summary, the experimental results show that the proposed Hybrid LSTM and Attention model (i.e., LSTM- SSAM) achieves the best accuracy and robustness with the least error over the various benchmarks and datasets.

## VI. CONCLUSIONS AND FUTURE WORK

For the stock market, learning the future price is very important for making investment decisions. This paper establishes a forecasting framework to predict the 'Adj Close' prices of stock. Compared to other approaches, we offer a novel stock market time series forecasting model called LSTM-SSAM, in which the self-attention model can more precisely analyze the sequential stock attributes. The experimental findings demonstrate that our proposed model has a lower RMSE and improved accuracy of the prediction outcomes when compared to the widely used time series models.

The fact that historical data alone cannot be used to reduce stock market volatility is still an open question for our future study in several directions, but other aspects of the current environment, such as the most recent news in politics and the economy, also need to be examined. In this paper, we have only focused on stocks related to the banking sector. Therefore, we can add data predictions related to stock news and Twitter sentiment analysis, so as to enhance the stability and accuracy of the model in the case of a major event.

We also envision the proposed approach could be made more practical by deploying it in a distributed fashion using recent ML paradigms such as distributed machine learning or federated learning approaches to provide for more dynamic and adaptive stock price prediction systems.

## VII. ACKNOWLEDGEMENT